# VĀCASPATI: A Diverse Corpus of Bangla Literature

**Pramit Bhattacharyya, Joydeep Mondal, Subhadip Maji, Arnab Bhattacharya**

Dept. of Computer Science and Engineering,
Indian Institute of Technology Kanpur
{pramitb,joydeep,subhadip,arnabb}@cse.iitk.ac.in

## Abstract

Bangla (or Bengali) is the fifth most spoken language globally; yet, the state-of-the-art NLP in Bangla is lagging for even simple tasks such as lemmatization, POS tagging, etc. This is partly due to lack of a varied quality corpus. To alleviate this need, we build VĀCASPATI, a diverse corpus of Bangla literature. The literary works are collected from various websites; only those works that are publicly available without copyright violations or restrictions are collected. We believe that published literature captures the features of a language much better than newspapers, blogs or social media posts which tend to follow only a certain literary pattern and, therefore, miss out on language variety. Our corpus VĀCASPATI is varied from multiple aspects, including type of composition, topic, author, time, space, etc. It contains more than 11 million sentences and 115 million words. We also built a word embedding model, VĀC-FT, using FastText from VĀCASPATI as well as trained an Electra model, VĀC-BERT, using the corpus. VĀC-BERT has far fewer parameters and requires only a fraction of resources compared to other state-of-the-art transformer models and yet performs either better or similar on various downstream tasks. On multiple downstream tasks, VĀC-FT outperforms other FastText-based models. We also demonstrate the efficacy of VĀCASPATI as a corpus by showing that similar models built from other corpora are not as effective. The models are available at https://bangla.iitk.ac.in/.

## 1 Introduction

Automated computational processing of natural language tasks has witnessed tremendous improvements in recent years, mostly due to availability of large text corpora and novel deep learning models to process those corpora. In case of English and some other western European languages where such corpora are available (e.g., the billion word corpus (Pomikálek et al., 2012)), state-of-the-art models in standard NLP tasks comprising of deep learning architectures outperform traditional rule-based models. Most other languages, however, do not enjoy such improved performances in NLP tasks due to a lack of large quality corpora.

Hence, in this paper, we build and release a large corpus, VĀCASPATI, for Bangla (Bengali, বাংলা, *bAMlA*)[1], which is the fifth most spoken language globally. It is the state language of several states in India including West Bengal and Tripura, besides being one of the official languages of India. Further, it is the main language for Bangladesh. Recently, there are proposals to adopt Bangla as one of the official languages in the UN (https://en.wikipedia.org/wiki/Official_languages_of_the_United_Nations).

There have been several attempts in the past to build a Bangla text corpus. Some of the notable ones are IndicCorp (Kakwani et al., 2020), that compiled a dataset for 11 Indian languages with news articles, including Bangla, and BanglaBERT (Bhattacharjee et al., 2022), that scraped data from social media posts, and ebooks from 110 sites. Sec. 2 details more related works.

One of the major concerns for all these corpora, however, is the quality and variety of the language. The sources mostly comprise of newspapers, blogs and social media posts. Newspapers are known to follow a certain style of language which is typically urban and devoid of common words such as "*mahIpati*" (king), "*hotRRigaNa*" (priest) used in day-to-day lives of native speakers. In addition, words such as "government" and "police" are unnaturally more frequent. Blogs and social media posts, on the other hand, are often full of grammatical mistakes, typos and non-native words and phrases (Kundu et al., 2013). Consequently, any language model built from such corpora are bound

---
[1] We have used "ITrans" format for transliteration of Bangla words (https://en.wikipedia.org/wiki/ITRANS).

to suffer from problems.

Thus, in this paper, we focus on only classical Bangla *literature* as our text source. We do not include any news article, blog or social media post. To the best of our knowledge, this is the first attempt to build a Bangla corpus using only literary works. BookCorpus (Zhu et al., 2015) with 11,038 books is a similar dataset in English.

We name our corpus VĀCASPATI which means "master of speech (or language)". VĀCASPATI is diverse from multiple aspects, including type of composition, topic, author, time, space, etc. (details in Sec. 3). VĀCASPATI consists of more than 11 million sentences and 115 million words, and is 2.1 GB in size.

Next, to examine if NLP tasks are getting facilitated by the corpus, we construct word embeddings using FastText, and build a BERT model. We test the effectiveness of these through several downstream tasks including word similarity task, poem and sentiment classification, and spelling error detection and correction. Notably, our models use only a fraction of the resources (running time, memory, number of parameters) as compared to the competing state-of-the-art models and yet either outperform them or give comparable results on these tasks.

In sum, our contributions in this paper are:
1. We create a large quality corpus, VĀCASPATI, using only Bangla literature (Sec. 3).
2. We construct word embeddings and build language models using VĀCASPATI (Sec. 4). Our models require 2-5 times less space and time as compared to competing ones.
3. Models built using VĀCASPATI either outperform or are similar to competing models on several downstream tasks (Sec. 5).

The models and test datasets are available at https://bangla.iitk.ac.in/projects/vacaspati.html.

## 2 Related Work

**Corpora:** Most Bangla datasets to date have been generated from newspaper articles. The EMILLE-CIIL (McEnery et al., 2000) monolingual written corpus consists of 1,980,000 words from newspapers, whereas the spoken corpus consists of only 442,000 words. Wikipedia dump of Bangla is of size 242MB (as on June 21, 2022). The Leipzig corpus (Goldhahn et al., 2012) on Bangla, curated by crawling newspapers, consists of 1,200,255 sentences and 16,632,554 tokens. The SUPara corpus (Mumin et al., 2012) is an English-Bangla sentence-aligned parallel corpus consisting of more than 200,000 words. OPUS (Tiedemann and Nygaard, 2004), a multilingual corpus of translated open-source documents, curated a sentence-aligned parallel corpus for Bangla with 4.7 million tokens and 0.51 million sentences. The BNLP (Sarker, 2021) dataset for embedding was curated by collecting 127,867 news articles and Wikipedia dumps, and was trained on 5.83 million sentences and 93.43 million tokens. Hasan et al. (2020b) built a sentence-aligned Bangla-English parallel corpus with 2.75M sentences focusing mainly on machine translation work. The IndicCorp corpus (Kakwani et al., 2020) for Bangla consists of 39.9 million sentences and 836 million tokens generated from 3.89 million news articles. The OSCAR project (Ortiz Suárez et al., 2019) built by crawling websites using CommonCrawl (https://commoncrawl.org) contains 632 million words for Bangla.

BanglaBERT (Bhattacharjee et al., 2022) generated a 27.5 GB Bangla dataset with 2.5 billion tokens by crawling 110 websites. BanglaBERT contains ebooks but also contains newspaper articles, blogs, and social media texts in addition to that. BanglaBERT only provides a pre-trained BERT model and does not provide any word embedding for their corpus. BERT is challenging to use for word-level tasks such as word similarity and word analogy as the words may not be present in vocabulary in their proper form.

**Word Embeddings**: Word embeddings have been trained on Bangla but either on limited data or only over news articles and social media blogs. The Polyglot (Al-Rfou' et al., 2013) project contains articles from Wikipedia and has only 55,000 words. Ahmad and Amin (2016) released a word embedding of 210,000 words created from newspaper articles. FastText (Bojanowski et al., 2017) provides embeddings trained either only on Wikipedia, or Wikipedia+CommonCrawl corpora (Grave et al., 2018). Bangla word embeddings are available from https://nlp.johnsnowlabs.com/2021/02/10/bengali_cc_300d_bn.html. IndicFT (Kakwani et al., 2020) provides embeddings trained on 3.9 million news articles and has 839 million tokens. Hossain and Hoque (2020) curated a dataset from 910,877 text files with 180 million words. VĀC-FT has 115 million words and is built on literary data, but still outperforms the other em-beddings.

| Time Period | #Words | #Sentences |
|---|---|---|
| 14-18th centuries | 1,865,284 | 290,036 |
| 19th century | 11,810,978 | 1,246,535 |
| 20th century | 80,583,660 | 8,050,493 |
| 21st century | 20,915,974 | 2,114,845 |
| Total | 115,176,214 | 11,701,910 |

Table 1: Temporal variation of VĀCASPATI

| Region | #Words | #Sentences |
|---|---|---|
| India | 75,805,789 | 7,689,915 |
| Bangladesh | 18,464,579 | 1,968,410 |
| Translations | 20,905,846 | 2,043,585 |
| Total | 115,176,214 | 11,701,910 |

Table 2: Spatial variation of VĀCASPATI

**Pretrained Transformers:** Transformers built on attention mechanism provided by (Vaswani et al., 2017) is a popular way of building language models and is useful in various downstream NLP tasks (Radford et al., 2019). Various transformer-based language models such as GPT (Brown et al., 2020), BERT (Devlin et al., 2018), RoBERTa (Liu et al., 2019), DistilBERT (Sanh et al., 2019), ALBERT (Lan et al., 2019), ELECTRA (Clark et al., 2020) have been proposed. For Indian languages, multilingual BERT such as XLM-R (Conneau et al., 2020), multilingual BERT (mBERT) (Pires et al., 2019), IndicBERT (Kakwani et al., 2020) and MuRIL (Khanuja et al., 2021) are available. BanglaBERT (Bhattacharjee et al., 2022), BanglishBERT (Bhattacharjee et al., 2022), sahajBERT (Diskin et al., 2021) are BERT models made specifically for Bangla. VĀC-BERT is lighter than all these variants (Table A5).

**Downstream Tasks:** The detailed related work on downstream tasks performed in this paper in Appendix A. Some of the notable works include (Rakshit et al., 2015) for poem classification, (Das and Bandyopadhyay, 2010; Hasan et al., 2020a) for sentiment classification, (Mandal and Hossain, 2017; UzZaman and Khan, 2005; Islam et al., 2018) for spelling correction, (Ekbal and Bandyopadhyay, 2008c,b, 2007; Chaudhuri and Bhattacharya, 2008; Chowdhury et al., 2018) for named entity recognition, etc.

## 3 The VĀCASPATI Corpus

Our aim is to generate a monolingual corpus for Bangla that can help build better and more accurate models for standard NLP tasks as well as in specialized downstream tasks (Jurafsky and Martin, 2021). Hence, we focused primarily on collecting literary data covering various types, including social, political, and religious. We scraped literature articles from various websites with publishers' permissions. Only the works of authors available on these websites are used for curating the dataset. More information regarding the ethical considerations are presented in Sec. 8.

Overall, the VĀCASPATI corpus consists of more than 115.1 million words and over 11.7 million sentences.

### 3.1 Variety in VĀCASPATI

The works collected in VĀCASPATI are written between 1310CE and the present year. The earliest works of Bangla language go back to this date and, thus, our corpus captures the temporal changes in the language features over time. One of the unique temporal features in Bangla is the transformation of sAdhu bhAShA (or *refined language*) to chalita bhAShA (or *colloquial language*). While till the 19th century, all written works were exclusively in *sAdhu bhAShA*, authors started switching to *chalita bhAShA* in different decades of the 20th century. Currently, almost all the works are in *chalita bhAShA*. The two differ mostly in verb forms and pronouns, and use exclusive sets of these. Consequently, a modern day native reader will find it hard to understand/read a piece of literature written in *sAdhu bhAShA* without practice. Newspaper articles, blogs and social media posts fail to capture this extremely unique transition phenomenon of the language. Table 1 shows the break-up of the different time periods.

In addition to the temporal aspect, Bangla has a lot of variety in language that differs from one region to another. Specific words and phrases are often used by authors from these regions. Hence, we ensured that our corpus contains works from authors of both sides of erstwhile Bengal (currently the West Bengal state in India and the country Bangladesh). This enables us to capture the spatial features of the language across regions that speak quite far apart dialects such as in Bankura in India and Chattogram in Bangladesh. Table 2 shows the spatial variations. In addition, it also shows the statistics for translated works.

VĀCASPATI consists of works of 6 prominent types. It includes poetry as well, since poems capture a different flavor of the language including words (such as "*mora*" meaning "my") that are exclusive to it. Table 3 shows the statistics of the types in our corpus. We collected essays as well.

| Type | #Words | #Sentences |
|---|---|---|
| Novels | 76,383,929 | 7,761,017 |
| Stories | 14,557,075 | 1,479,077 |
| Poetry | 2,939,392 | 298,658 |
| Drama | 1,263,240 | 128,352 |
| Letters | 145,366 | 14,770 |
| Essays | 19,887,212 | 2,020,036 |
| Total | 115,176,214 | 11,701,910 |

Table 3: Type of works in VĀCASPATI

The works collected in VĀCASPATI are also arranged according to topics (Table 4). Notably, we have specialized categories such as children, law and religion, most of which are generally completely absent in newspapers and are rare even in blogs and social media posts. This highlights the variety of our corpus.

### 3.2 Data Cleaning and Pre-Processing

Since VĀCASPATI comprises literary texts from the 14th century, usage of old punctuation marks are prevalent in the corpus. For example, "।...", "॥" were prevalent in the 16th century and earlier texts, but are extinct now. Due to this issue, tokenization modules such as iNLTK (https://inltk.readthedocs.io/en/latest/api_docs.html) fail to capture these punctuation marks and consider them as part of words. iNLTK fails to capture many such characters and words, and introduces (arbitrary) Unicode characters in their place.

Appendix B shows a list of such punctuation marks and Unicode characters. After cleaning all such non-Bangla Unicode characters and old punctuation marks, we split the text based on whitespaces to generate words.

## 4 Embeddings and Models

We now describe the various word embedding and other models built using VĀCASPATI.

### 4.1 VĀC-FT

Bangla is morphologically rich and is invested with inflections. Hence, we start with FastText that can integrate sub-word information using n-gram embeddings during training.

We train FastText embeddings for Bangla using VĀCASPATI and evaluate their quality on two tasks, word similarity and spell-checking. We train 115M+ words generated from VĀCASPATI corpus on a 300-dimensional vector space using FastText with Gensim (https://pypi.org/project/gensim/). Our skip-gram models have been trained for 100 iterations with a window size of 10 and a minimum word size of 4 characters. For each instance, 10 negative examples were used. These parameters are chosen based on suggestions by (Grave et al., 2018) and based on the average length of words and characters in words of a Bangla sentence (9.84 words per sentence and 5.36 characters per word) present in VĀCASPATI.

| Topic | #Words | #Sentences |
|---|---|---|
| Agriculture | 873,972 | 77,497 |
| Children | 3,847,634 | 383,719 |
| Comedy | 2,504,177 | 278,263 |
| Economy | 248,379 | 16,542 |
| History | 11,630,535 | 1,199,054 |
| Law | 104,361 | 2,590 |
| Nationalism | 387,214 | 27,658 |
| Philosophy | 667,541 | 42,584 |
| Political | 120,421 | 5,421 |
| Religion | 7,560,964 | 804,380 |
| Science Fiction | 2,466,577 | 131,172 |
| Scientific | 999,298 | 65,847 |
| Social | 71,852,698 | 7,220,852 |
| Sports | 553,632 | 50,404 |
| Thriller | 8,830,192 | 1,114,970 |
| Travelogue | 2,528,619 | 280,957 |
| Total | 115,176,214 | 11,701,910 |

Table 4: Distribution of topics in VĀCASPATI

Since (Kumar et al., 2020) showed that for inflectional languages such as Bangla, FastText performs better than Glove (Pennington et al., 2014) and Word2Vec (Mikolov et al., 2013), hence, we show the results for only our FastText word vectors. We name this VĀC-FT.

### 4.2 VĀC-BERT

We next introduce VĀC-BERT, which is built from VĀCASPATI by pre-training using Electra, with the Replaced Token Detection (RTD) objective. A sequence with 15% as masked tokens is fed to the generator, which predicts the rest of the input. After replacing masked tokens with the generator's output distribution, the discriminator must predict whether each token is from the original sequence. The discriminator is used for fine-tuning. Since Electra shows comparable performance (Clark et al., 2020) in several downstream tasks as compared to larger models such as RoBERTa (Liu et al., 2019) and XLNet (Yang et al., 2019) with only a fraction of their training time and memory, we used Electra for our implementation of VĀC-BERT. We next describe in detail the pre-training and fine-tuning steps.

### 4.2.1 Pre-training

Using VĀCASPATI, we first train a Word Piece tokenizer (Wu et al., 2016) using n-gram embeddings to tokenize the sentences. We use this tokenized corpus to train VĀC-BERT. We pre-trained the small variant of Electra model (a 12-layer transformer encoder having an embedding dimensionality of 128, hidden layer size of 256, 4 attention heads, feed-forward size of 3072, generator-to-discriminator ratio 1:3, and 50,000 vocabulary size, amounting to a total of 16.8 million parameters) with 64 batch sizes for 250K steps on a 40 GB instance of NVidia A100 GPU. We used the Adam optimizer (Kingma and Ba, 2015) with a 2e-4 learning rate. Training time was <12 hours.

Our model is 7 times lighter than BanglaBERT (Bhattacharjee et al., 2022), which has 110M parameters and is trained for 2.5M steps in a TPU v3.0 instance. IndicBERT (Kakwani et al., 2020), on the other hand, has 18M parameters and is trained for 400k steps on a TPU v3.0 instance. Other than these, we have also compared the performance of our VĀC-BERT with a few other standard BERT models, including XLM-R base (280M parameters) (Conneau et al., 2020), XLM-R large (550M parameters), sahajBERT (18M parameters) (Diskin et al., 2021) mBERT (180M parameters) (Pires et al., 2019), and MuRIL (236M parameters) (Khanuja et al., 2021). Thus, compared to all the state-of-the-art BERT models for Bangla, our VĀC-BERT model is the lightest, and fastest to build.

### 4.2.2 Fine-tuning

After pre-training, we fine-tune VĀC-BERT on each task using the respective training sets. The fine-tuning is done independently for each task (i.e., we have a task-specific model) as described next in Section 5.

## 5 Evaluation

In this section, we quantitatively evaluate our corpus VĀCASPATI through various downstream tasks. We test both simple word embeddings as well as transformer-based models.

### 5.1 Word Embedding

We benchmark VĀC-FT word embeddings against three FastText embeddings: (1) IndicFT, trained on the Bangla subset of IndicCorp (Kakwani et al., 2020), and pre-trained embeddings released by the FastText project trained on (2) Wikipedia, denoted by FT-W (Bojanowski et al., 2017), and (3) Wiki+CommonCrawl, denoted by FT-WC (Grave et al., 2018).

| Model | Word Similarity | Poem (5-class) | Sentiment (2-class) | Sentiment (3-class) | Average Macro-F1 |
|---|---|---|---|---|---|
| Indic-FT | 57.00% | 41.8% | 46.23% | 45.19% | 47.24% |
| FT-W | 56.00% | 40.5% | 44.54% | 44.15% | 46.00% |
| Ft-WC | 55.00% | 39.2% | 43.68% | 43.35% | 45.00% |
| VĀC-FT | **64.50%** | **47.5%** | **47.35%** | **45.30%** | **50.74%** |

Table 5: Macro-F1 of different FastText models.

We evaluate the FastText word embeddings in two classes of tasks: (1) word similarity, and (2) classification-based. Unless otherwise mentioned, we evaluate the *macro-F1 score* throughout to compare the methods. We choose macro-F1 since the data classes are unbalanced.

#### 5.1.1 Word Similarity

Bangla suffers from a lack of quality test sets for the word similarity task. IIIT-Hyderabad released a dataset (Akhtar et al., 2017) for similarity measures of 100-200 word pairs for seven languages that do not include Bangla. We created a *benchmark dataset* for Bangla with 600 word pairs to fill the gap. These words are chosen directly from a well-known Bangla grammar book (Chattopadhyay, 1942) and are classified in 4 categories similar to the one specified in the book: *synonyms*, *antonyms*, *homonyms*, and *dissimilar*. The details are in Appendix C.1.

The word similarity task has been treated as a *binary classification problem* where a cosine similarity value between the two word embeddings above a threshold is considered as a "similar" class; otherwise, the class is "dissimilar". While synonyms form the "similar" class of word pairs, the other three constitute the "dissimilar" class. We kept the cosine similarity threshold as 0.5 after running experiments at intervals of 0.05 from 0.2 to 0.9.

Table 5 shows that the macro-F1 score of VĀC-FT is the best (64.5%), and is better than others by 7-9%. The performance of VĀC-FT on each class of words, along with its analysis, is described in Appendix C.2 (Table A1).

#### 5.1.2 Classification Tasks

We next evaluate the FastText embeddings on different text classification tasks:

- **Poem Classification:** We evaluate a subject-based classification task on poems written by Rabindranath Tagore, who is one of the greatest poets of Bangla, into 5 categories as indi-cated by the author himself. The categories are *pUjA* (devotion), *prema* (love), *prAkRRiti* (nature), *svadesha* (nationalism), and *vichitrA* (miscellaneous). We collected the poems from the website of The Complete Works of Tagore available at https://tagoreweb.in/. Five-fold cross-validation was done on all 1,451 poems.

- **Sentiment Classification:** Sentiment analysis is the task of classifying people's opinions and emotions towards entities such as products, services, organizations, and others. Sentiment classification is a prevalent downstream task in Bangla to indicate the efficacy of the corpus. Some of the earlier works of sentiment classification were done a decade back (Das and Bandyopadhyay, 2010). For our work, we use two publicly available datasets. The first dataset (Sazzed, 2020) consists of 3,307 negative and 8,500 positive reviews annotated on YouTube Bangla drama. The second dataset (Islam et al., 2021) comprises 3 polarity labels, positive, negative, and neutral, and is collected from social media comments on news and videos covering 13 domains, including politics, education, and agriculture. It consists of 5,709 negative, 6,410 positive, and 3,609 neutral sentences.

### 5.1.3 Results on Classification Tasks

Since we wanted to evaluate the effect of FastText embeddings built from the corpus on the classification tasks, we chose *k-nearest-neighbors (kNN)* as our classifier following Meng et al. (2019) who argued that the classification performance of a non-parametric classifier indicates the best how well text semantics have been captured by the FastText embeddings. The input text embedding is the mean of an article's word embeddings.

The results of all the classification tasks is shown in Table 5 (the best results for all the FastText models were obtained for k=11). VĀC-FT performs the best for all the classification tasks. This shows that VĀC-FT captures the semantics of words better, and is effective for even tasks such as sentiment classification that does not use literary data. For literary data such as poems, it is significantly better.

## 5.2 Transformer-based Models

After establishing the superior performance of our corpus for word embedding based tasks, we switch to state-of-the-art transformer models and test the performance on various downstream tasks.

### 5.2.1 Tasks

In addition to *poem classification* (5-class) and *sentiment classification* (both 2-class and 3-class) tasks as described previously in Section 5.1.2, we conduct two more tasks that are possible with transformer-based models:

- **Spelling error detection:** Spelling error is a customary problem in every language. The prevalent Bangla spell-checkers (Mandal and Hossain, 2017; UzZaman and Khan, 2005) do not take into account the context and, therefore, fail to deal with many real-world spelling errors such as "বিষ" (*viSha*) versus "বিশ" (*visha*) and "কোণ" (*koNa*) versus "কোন" (*kona*) since both the words are present in Bangla vocabulary. The issues get greatly manifested in texts produced through OCR, such as when "সূর্য অস্ত গেল" (*sUrya asta gela*, "the sun set") gets changed to "সূর্য অন্ত গেল" (*sUrya anta gela*, "the sun ended"). In both these sentences, all the words are correct and are part of Bangla vocabulary. To correct such errors, the model needs to understand the semantics and context of a sentence.

  To the best of our knowledge, there is no publicly available dataset for context-sensitive spelling error detection in Bangla. Hence, we created our own dataset using OCR. Appendix D.1 mentions the detailed procedure of dataset creation. We generated a dataset of 110,356 sequence pairs.

  We used VĀC-BERT to detect errors in the sentence. We treated this problem as a sequence pair classification task where a sentence is passed as one sequence, and every word in it is passed as the second sequence to determine whether it is contextually correct. The dataset is divided into 80%-20% for training and testing.

- **Named Entity Recognition (NER):** NER (Named Entity Recognition) is a sequence labeling task that finds spans of text and tags them to a named entity class (Jurafsky and Martin, 2021). We feed each sentence to the model as a single sequence. For every token, a softmax layer computes the probability distribution over the classes. Multi-class cross-entropy loss is used for fine-tuning the model.

  We chose the publicly available Wikiann (Bangla subset) (Pan et al., 2017) NER dataset created from Wikipedia data. The dataset consists of 15,445 sentences (12,356 sentences used for training and 3,089 sentences used for test-ing) and more than 1,35,000 tokens. The tokens are tagged into 4 classes: person (Per), location (Loc), organization (Org), other (O).

Thus, of the 5 tasks that we test, poem classification and spelling error detection can be considered more as *literature tasks* in the sense that a corpus with literature data is likely to do better, while the other three are *non-literature tasks*.

## 5.3 Performance of VĀC-BERT

We evaluated the performance of our transformer-based model, VĀC-BERT, built on our VĀCASPATI corpus vis-a-vis other state-of-the-art transformer-based models available in Bangla on the tasks specified above. The models include *monolingual* models such as BanglaBERT and sahajBERT as well as *multilingual models* such as XLM (base and large), mBERT, IndicBERT, MuRIL, and BanglishBERT. The details of these models are discussed in Section 2. Unless otherwise mentioned, the *macro-F1 score* is used as the evaluation metric for all the tasks.

### 5.3.1 Results

Table 6 shows the performance of VĀC-BERT and other transformer-based models on different tasks.

The macro-F1 score of our VĀC-BERT model is 60.39% for poem classification, which is the best.

We have also performed the best for sentiment classification (3 class) and is within 0.25% of macro-F1 score of the best competing model (MuRIL and BanglaBERT) in sentiment classification (2 class). The task of sentiment classification is performed on public non-literary datasets.

In the spelling error detection task, we are within ∼1% of BanglaBERT. An in-depth analysis, however, shows that VĀC-BERT gives 55.0% recall on the error class (i.e., wrongly spelled words) as compared to 52.0% for BanglaBERT (Table A3 in Appendix D.2). This implies that VĀC-BERT is able to detect more context-sensitive spelling errors. Detecting more errors gives a better chance of correcting errors in the next phase. Thus, the lower precision in the error class can be improved in the error correction stage later.

Our model performs comparably to the best models (within ∼2%) on the NER task performed on a public dataset taken from Wikipedia. Hence, BanglaBERT and MuRIL (who are better than VĀC-BERT) may have already seen this data.

### 5.3.2 Corpus Efficacy

In the previous section, we established that models built on VĀCASPATI with fewer parameters either perform better or are comparable to state-of-the-art models. In this section, we will show that VĀCASPATI, as a *corpus*, has telling effects on the results obtained for the downstream tasks and, thus, it is not only the models due to which gains are observed.

To establish that, we pre-trained an Electra-small transformer, Indic-ELECTRA, with the *same hyper-parameters and architecture* as VĀC-BERT, on the Bangla subset of IndicCorp (Kakwani et al., 2020). We could not do this for the other models since their corresponding corpora are not available publicly. We also *combine* (i.e., take union with) the Bangla subset of the IndicCorp corpus with our VĀCASPATI corpus, and pre-train another Electra-small transformer with again the same hyper-parameters and architecture. We call this model IV-ELECTRA.

Table 6 shows that VĀC-BERT outperforms IndicElectra on all the tasks (the last 2 rows above VĀC-BERT, which is shown separately as Table A4 in Appendix F). Since the only change between the two models is the corpus (VĀCASPATI versus IndicCorp), this indicates the possible superiority of VĀCASPATI as a *corpus*. VĀC-BERT also outperforms IV-ELECTRA on average, even on non-literary datasets. It indicates that models built from only VĀCASPATI (literary corpus) are good enough to perform language tasks. Further, the requirement for a large dataset, which is cumbersome to curate for a low-resource language like Bangla, is also done away with.

Table 6 captures one more interesting effect. The performance of Indic-ELECTRA is enhanced on adding VĀCASPATI to the IndicCorp dataset much more than is done for VĀCASPATI by adding IndicCorp to it. Thus, the incremental benefit in performance of models by adding a literature dataset over a non-literature dataset is much more significant than that by adding a non-literature dataset over a literature dataset. In fact, in the literature tasks, it reduces the performance. This is due to poorer quality of language as compared to a literature corpus and presence of various types of errors. This reinforces the importance of quality data in a corpus.

| Model | Parameters | Literature | | | | Non-literature | | | | | Overall | |
|---|---|---|---|---|---|---|---|---|---|---|---|---|
| | | Poem (5-class) | Spell Error Detection | Literature Average | Δ Literature | Sentiment (2-class) | Sentiment (3-class) | NER (4-class) | Non-Literature Average | Δ Non-Literature | Overall Average | Δ Overall |
| mBERT | 180M | 59.85 | 41.00 | 50.43 | -18.63 | 94.80 | 63.68 | 86.51 | 81.66 | -2.75 | 69.17 | -9.10 |
| XLM-R (base) | 270M | 35.00 | 42.50 | 38.75 | -33.12 | 94.64 | 67.8 | 87.50 | 83.31 | -2.20 | 65.50 | -14.47 |
| XLM-R (large) | 550M | 31.86 | 40.00 | 35.93 | -30.30 | 93.89 | 65.75 | 87.60 | 82.21 | -1.10 | 63.80 | -12.77 |
| IndicBERT | 18M | 43.67 | 41.50 | 42.59 | -26.46 | 91.86 | 59.34 | 88.31 | 79.84 | -4.57 | 64.94 | -13.33 |
| MuRIL | 236M | 52.36 | 42.50 | 47.43 | -21.62 | **95.00** | 68.73 | 90.62 | 84.78 | +0.37 | 69.84 | -8.43 |
| BanglishBERT | 110M | 49.86 | 74.56 | 62.21 | -6.84 | 93.45 | 66.58 | 89.52 | 83.18 | -1.23 | 74.79 | -3.48 |
| BanglaBERT | 110M | 54.40 | **78.90** | 66.65 | -2.40 | **95.00** | 68.68 | **91.52** | **85.06** | +0.65 | 77.72 | -0.55 |
| sahajBERT | 18M | 46.34 | 70.89 | 58.61 | -10.44 | 93.75 | 67.52 | 85.50 | 82.25 | -2.16 | 72.80 | -5.47 |
| Indic-ELECTRA | 17M | 48.46 | 74.58 | 61.52 | -7.53 | 93.86 | 67.2 | 89.08 | 83.38 | -1.03 | 74.63 | -3.64 |
| IV-ELECTRA | 17M | 60.00 | 76.64 | 68.32 | -0.73 | 94.25 | 66.31 | 90.30 | 83.62 | -0.79 | 77.50 | -0.77 |
| **VĀC-BERT** | 17M | **60.39** | 77.72 | **69.05** | **0.00** | 94.75 | **68.79** | 89.70 | 84.41 | **0.00** | **78.27** | **0.00** |

Table 6: Parameters and performance of different transformer-based models (performance measure is macro-F1).

| Time-Period | Poem (5-class) | Sentiment (2-class) | Sentiment (3-class) |
|---|---|---|---|
| Pre-1941 VĀC-BERT | 63.89 | 92.67 | 60.89 |
| Post-1941 VĀC-BERT | 56.56 | **94.75** | 68.95 |
| Complete VĀC-BERT | **60.39** | **94.75** | 68.79 |

Table 7: Performance on temporal variations

## 5.4 Necessity of Temporal Variation

To assess the necessity of inclusion of works from different temporal periods, we segregated VĀCASPATI into two parts, *pre-1941* and *post-1941*. The most influential litterateur of Bangla, namely Rabindranath Tagore, passed away in 1941, which had a large effect on the Bangla literary style. We pre-train the same Electra model (same hyper-parameters and architecture) as used in VĀC-BERT on both divisions of the dataset.

Table 7 shows the performance of the split corpora on 3 downstream classification tasks. Pre-1941 data significantly outperformed post-1941 data on the poem classification task, whereas for sentiment classification tasks, the post-1941 data performed better. The complete VĀCASPATI corpus performed equally well on both the tasks.

The poem classification data was based on Rabindranath Tagore's poetry, which shows that a corpus with only modern writings may not have the literary variety to capture nuances of earlier works. This also shows why corpora built from newspapers, blogs and social media posts will perform poorly on older literary data. The test set for sentiment classification task is on modern writing. Although the pre-1941 data performs fairly, having modern examples of the language helps to improve the performance. VĀCASPATI, which includes both pre-1941 as well as post-1941 data achieves the best balance.

## 6 Discussions and Future Work

In this paper, we proposed a quality Bangla corpus based only on literature, called VĀCASPATI, that includes variations across several aspects.

The experiments showed that VĀCASPATI, as a corpus, has better quality than other corpora, since it could outperform models built on those corpora using the same hyper-parameters and architecture. Also, the temporal variation in VĀCASPATI is useful since it provides a nice balance between downstream tasks on modern Bangla and older variants.

Further, VĀC-FT model built using VĀCASPATI has far fewer out-of-vocabulary words as compared to other FastText models and, thus, outperforms them on the word similarity task. VĀC-FT also outperforms other models on all the classification tasks which also establishes the quality of VĀCASPATI as a corpus.

VĀC-BERT, which is an Electra-small variant built from VĀCASPATI, has either better or similar performance to other BERT models on various downstream tasks. On average, VĀC-BERT outperforms all the competing BERT models. Since VĀCASPATI is a corpus made from only literary data, VĀC-BERT did not use newspaper or social media articles during pre-training. Most of the other BERT models had most likely used those data during pre-training. Hence, the performance of VĀC-BERT on non-literary data is significant to establish the quality of VĀCASPATI as a corpus.

The experiments with (the Bangla subset of) IndicCorp using the exact same model architecture and hyper-parameters shows that VĀCASPATI is a better corpus. Also, addition of non-literary data on top of literary data produces minimal incremental improvement while the reverse can produce significant benefits.

VĀC-BERT was the fastest and it used the least amount of memory. This makes it easiest to deploy for real-world applications.

**Future Work**: A quality corpus for any language is useful to get better performing models for various NLP tasks including low-level tasks such as lemmatization, POS tagging, NER, dependency parsing, etc. We hope our VĀCASPATI corpus will fuel more research in building better models for the Bangla language.

# 7 Limitations

Curating a quality dataset from literary articles for Bangla is challenging as not many books are available as text files without copyright issues. OCR texts are noisy and are, thus, not included in the dataset. FastText embeddings and BERT models may improve with more data; however, training larger models is hindered by resource constraints. More data may not always reflect a gain the performance, since it may be noisy.

# 8 Ethics Statement

The VĀCASPATI corpus dataset has been curated by scraping literary works available in the public domain. They are taken from websites that have the authors' permission to showcase their works. We restricted our dataset to only such works. Hence, no copyright infringement has been done. We will release the VĀC-BERT model and the VĀC-FT word embedding vectors under a non-commercial license upon acceptance of the paper. The datasets and code for all the downstream tasks will also be made publicly available under a similar non-commercial license. Since the corpus is very large, it was not feasible to take steps to detect offensive content. However, since we only used published literary works, it is unlikely that our dataset contains highly objectionable content. Other than these, we see no other ethical concerns in the paper.

## A    Related Work on Downstream Tasks

Poem classification into subject-based categories or genres is a challenging downstream task as it requires understanding the semantics and nuances of the language. Rakshit et al. (2015) tried the same on Rabindranath Tagore's works by segregating poems into 4 categories. In this paper, we include the more challenging miscellaneous category to make it a 5-class problem.

Das and Bandyopadhyay (2010) has evaluated the polarity of opinionated phrases on news articles using SVM into two classes, positive and negative. Hasan et al. (2020a) has, however, shown that transformer-based models outperform other models for Bangla.

The prevalent Bangla spell checkers are mostly context-free. Mandal and Hossain (2017) used a clustering-based edit distance model while UzZaman and Khan (2005) used a mapping rule-based edit distance and double metaphones to develop an automatic Bangla spell-checker. Islam et al. (2018) developed an N-gram model with edit distance for automatically correcting misspelled words. Unlike other works, our work considers the context of the sentence.

NER (Named Entity Recognition) is a sequence labelling task that finds spans of text constituting proper names and tags them to a named entity (Jurafsky and Martin, 2021). Most of the NER works in Bangla are conducted a decade ago. Ekbal et al. focused on corpus development (Ekbal and Bandyopadhyay, 2008c) (Ekbal and Bandyopadhyay, 2008b) feature engineering and using HMMs (Ekbal and Bandyopadhyay, 2007), CRF (Ekbal et al., 2008), SVMs (Ekbal and Bandyopadhyay, 2008a), and ME (Multi-Engine) (Ekbal and Bandyopadhyay, 2009) for the NER task. The reported Micro F1 varies between 82% to 91%. for various entities. Chaudhuri and Bhattacharya (2008) uses a hybrid approach using rule and n-gram based statistical modeling. Chowdhury et al. (2018) developed LSTM with CRF model and achieved a Micro F1 score of 72%.

## B    Data Cleaning

- *Cleaning of Unicode characters*: Unicode characters "0020" (space), "00a0" (no-break space), "200c" (zero width non-joiner), "1680" (ogham space mark), "180e" (mongolian vowel separator), "202f" (narrow no-break space), "205f" (medium mathematical space), "3000" (ideo-graphic space), "2000" (en quad), "200a" (hair space) are separated from the texts.

Table A1: FastText models for word similarity task

| Model | Similar | | | Dissimilar | | | Macro-F1 |
|---|---|---|---|---|---|---|---|
| | P | R | F1 | P | R | F1 | |
| IndicFT | **71%** | 43% | 54% | 50% | **75%** | **60%** | 57.0% |
| FT-W | 69% | 42% | 52% | 50% | 74% | **60%** | 56.0% |
| FT-WC | 69% | 41% | 51% | 50% | 73% | 59% | 55.0% |
| VĀC-FT | 70% | **71%** | **70%** | **60%** | 59% | 59% | **64.5%** |

Table A2: VĀC-BERT on 5-class poem classification

| Category | #Poems | Precision | Recall | F1 |
|---|---|---|---|---|
| *pUjA* | 617 | 63.0% | 61.0% | 61.98% |
| *prema* | 395 | 80.0% | 80.0% | 80.00% |
| *prAkRRiti* | 283 | 67.0% | 76.0% | 71.21% |
| *svadesha* | 46 | 80.0% | 44.0% | 56.77% |
| *vichitrA* | 110 | 32.0% | 32.0% | 32.00% |
| Total | 1451 | 64.4% | 58.6% | 60.39% |

Table A3: Performance on spelling error detection.

| Model | Correct Sentence | | | Incorrect Sentence | | |
|---|---|---|---|---|---|---|
| | P | R | F1 | P | R | F1 |
| BanglaBERT | 87% | 98% | 92% | 89% | 52% | 66% |
| VĀC-BERT | 88% | 95% | 91% | 77% | **55%** | 64% |

- *Cleaning of different punctuation marks*: In Bangla, usage of punctuation marks has also evolved alongside words. In particular, we have treated the following as punctuation marks: "...", "ι...", " ι ι", "!–", "–".

## C Word Similarity

### C.1 Details of Categories

The details of the 4 categories are:

- *Synonyms*: There are 347 synonymous word pairs such as "*mAtA*" and "*jananI*", which mean "mother", in the test dataset.
- *Antonyms*: These word pairs (119 in number) are opposite to each other, e.g., "*duShTa*" (bad) and "*shiShTa*" (good).
- *Homonyms*: These word pairs (84 in number) sound the same, but their meanings are different, e.g., "*kona*" (which) and "*koNa*" (angle).
- *Dissimilar*: These words are completely dissimilar, e.g., "*nadI*" (river) versus "*shIta*" (cold). There are 50 such word pairs.

### C.2 Details and Analysis of Results

Table A1 shows the class-wise precision, recall and F1 scores for both similar and dissimilar pairs of words.

VĀC-FT correctly classifies synonymous word pairs such as "*rAjA*" (king) and "*mahIpati*" (king); other embeddings failed to capture the word "*mahIpati*" at all due to lack of variety in their corpus. For word pairs such as "*udvRRitta*" (excess) and "*ghATati*" (dearth) that are antonyms of each other, only VĀC-FT gives the correct result. Since antonyms are semantically opposite of each other, they often occur in similar context in texts and, thus, are predicted to have similar FastText embeddings. Also, in Bangla, antonyms often differ in only one character ("*sakarmaka*" or "transitive verb" versus "*akarmaka*" or "intransitive verb"), which is hard for models such as FastText to capture.

## D Spelling Error Detection

### D.1 Dataset Creation

We took two books for which both pdf copies as well as the actual text (through scraping) are available freely. The pdf versions were subjected to OCR using Tesseract (https://pypi.org/project/pytesseract/). The corresponding text data were considered as the ground truth. This spell-checking dataset was *not present* during the pre-training of VĀC-BERT.

Since OCR consists of fragmented and missed sentences, we map 30-length word sequences between OCR data and scraped data, and map the pair of words with highest similarity in the sentence using edit distance. The threshold for similarity is kept at 0.6 which was chosen empirically after considering all the thresholds between 0.1 and 1.0 at intervals of 0.05.

### D.2 Results

Table A3 shows the performance of BanglaBERT and VĀC-BERT on the spelling error detection task. VĀC-BERT has a higher recall for the error class, which means it can detect more context-sensitive spelling errors. Detecting more errors gives a better chance of correcting errors in the next phase.

## E Poem Classification

Table A2 shows the details.

## F Corpus Efficacy

Table A4 shows the effect of changing the corpus using the same transformer model with the same hyper-parameters and architecture.

| Model | Parameters | Literature | | | | Non-literature | | | | | Overall | |
|---|---|---|---|---|---|---|---|---|---|---|---|---|
| | | Poem (5-class) | Spell Error Detection | Literature Average | Δ Literature | Sentiment (2-class) | Sentiment (3-class) | NER (4-class) | Non-Literature Average | Δ Non-Literature | Overall Average | Δ Overall |
| Indic-ELECTRA | 17M | 48.46 | 74.58 | 61.52 | -7.53 | 93.86 | 67.2 | 89.08 | 83.38 | -1.03 | 74.63 | -3.64 |
| IV-ELECTRA | 17M | 60.00 | 76.64 | 68.32 | -0.73 | 94.25 | 66.31 | 90.3 | 83.62 | -0.79 | 77.50 | -0.77 |
| **VĀC-BERT** | 17M | **60.39** | **77.72** | **69.05** | **0.0** | **94.75** | **68.79** | 89.70 | **84.41** | **0.0** | **78.27** | **0.0** |

Table A4: Comparison with IndicElectra.

| Model | Max Fine-tuning Time (secs/epoch) | Max Memory Usage (GB) |
|---|---|---|
| IndicBERT | 9.74 | 14.30 |
| MuRIL | 38.66 | 24.45 |
| BanglaBERT | 43.09 | 25.65 |
| BanglishBERT | 39.35 | 25.65 |
| sahajBERT | 46.35 | 27.15 |
| mBERT | 44.45 | 25.73 |
| XLM-R (base) | 44.95 | 25.74 |
| XLM-R (large) | 52.35 | 34.63 |
| VĀC-BERT | **3.46** | **10.60** |

Table A5: Time and space usage of transformer-based models for Bangla.

## G Efficiency of Transformer Models

We monitored the training time and memory usage during the fine-tuning stage for downstream tasks for the transformer-based models for Bangla including VĀC-BERT. The results are shown in Table A5. All the tasks were run on an A100 GPU machine with 40 GB memory. All the models had the same batch size (16) and sequence length (128) for comparisons.